\documentclass[letterpaper, 10 pt, conference]{ieeeconf}  % Comment this line out if you need a4paper

\IEEEoverridecommandlockouts                              % This command is only needed if 
\usepackage{amssymb, amsmath}
\DeclareMathOperator*{\argmax}{arg\,max}
\DeclareMathOperator*{\argmin}{arg\,min}

\usepackage{graphicx}
\usepackage{wrapfig} % wrap the text around a figure
\usepackage{subcaption} % insert several images in one figure
\usepackage{algorithm}
\usepackage{algorithmic}
\usepackage{cite}
\usepackage{tabularx}
\usepackage{array}
\newcolumntype{Y}{>{\centering\arraybackslash}X}

\title{\LARGE \bf
FPAS: Frontier-Based Path Planning with Adaptive Sampling for Large-Scale Unknown Environments
}

\author{Jinwoo Choi, Yeonkyu Lee, Jung-Taak Kim, Jisung Bae, Seung-Woo Seo% <-this % stops a space
\thanks{This research was supported by the Challengeable Future Defense Technology Research and Development Program through the Agency for Defense Development (ADD) funded by the Defense Acquisition Program Administration (DAPA) in 2026 (No. 915108201). (\textit{Corresponding author: Seung-Woo Seo})}
\thanks{The authors are with the Department of Electrical and Computer Engineering, Seoul National University, Seoul, South Korea (email: \{wlsdn9350, yeonkyu011, mychoco333, trr0631, sseo\}@snu.ac.kr)}%
}

\begin{document}

\maketitle
\thispagestyle{empty}
\pagestyle{empty}

%%%%%%%%%%%%%%%%%%%%%%%%%%%%%%%%%%%%%%%%%%%%%%%%%%%%%%%%%%%%%%%%%%%%%%%%%%%%%%%%
\begin{abstract}

In this work, we propose Frontier-based Path Planning with Adaptive Sampling (FPAS), a novel framework designed for efficient goal-reaching in large-scale, unknown environments. While existing planners often struggle with computational bottlenecks or inefficient paths during long-range navigation, FPAS overcomes these challenges by reinterpreting the frontier concept for goal-directed tasks. Specifically, our method leverages frontiers to effectively guide forward progression into unobserved regions and to select promising subgoals for backtracking from dead-ends or inefficient paths. Furthermore, FPAS introduces an adaptive sampling mechanism based on a frontier-derived openness metric. This mechanism dynamically adjusts the global graph's density by employing sparse nodes in open areas to alleviate computational burdens, while preserving denser sampling in narrow passages to ensure connectivity. Extensive evaluations demonstrate that FPAS substantially improves computational efficiency over baseline methods while maintaining highly competitive goal-reaching performance.

\end{abstract}

%%%%%%%%%%%%%%%%%%%%%%%%%%%%%%%%%%%%%%%%%%%%%%%%%%%%%%%%%%%%%%%%%%%%%%%%%%%%%%%%
\section{Introduction}

Generating paths to reach goals in large-scale unknown environments is a challenging task that has become increasingly relevant in various domains, including autonomous driving\cite{dolgov2010path}, robotics\cite{yang2016survey}, search and rescue missions\cite{delmerico2019current}, and military operations\cite{roberge2018fast}. The primary objective is to find a goal-directed path that minimizes travel cost and time under limited observations, while remaining robust to uncertainty in the environment. Because the map is only partially known, the planner must continuously adapt its decisions to newly revealed space while efficiently managing the growing volume of accumulated spatial data.

Despite active research, existing path generation approaches often struggle to meet these demands and fall short when reaching goals in large-scale unknown environments. These methods can be broadly categorized into search-based, sampling-based, and visibility-graph-based approaches. While search-based methods (e.g., A*\cite{hart1968formal}, D*\cite{stentz1994optimal}) offer strong optimality, their search costs scale poorly when frequent plan revisions and continuous map updates are required. Sampling-based methods like RRT are highly applicable in continuous spaces. However, their inherent randomness and goal-directed expansion heuristics often cause unnecessary detours or back-and-forth behavior in complex terrains. Recently, visibility-graph-based methods, notably the FAR Planner\cite{yang2022far}, have emerged as powerful alternatives in unknown environments. Yet, as the number of vertices increases, the cost of checking visibility connections and updating the graph escalates quickly. Consequently, in scenarios where the graph expands due to complex obstacle distributions or increased exploration time, computational overhead remains a significant bottleneck. Therefore, a novel framework is critical to appropriately balance goal-reaching with the exploration of unobserved regions, while efficiently managing the memory of environmental information.

\begin{figure}
    \centering
    \includegraphics[width=0.48\textwidth]{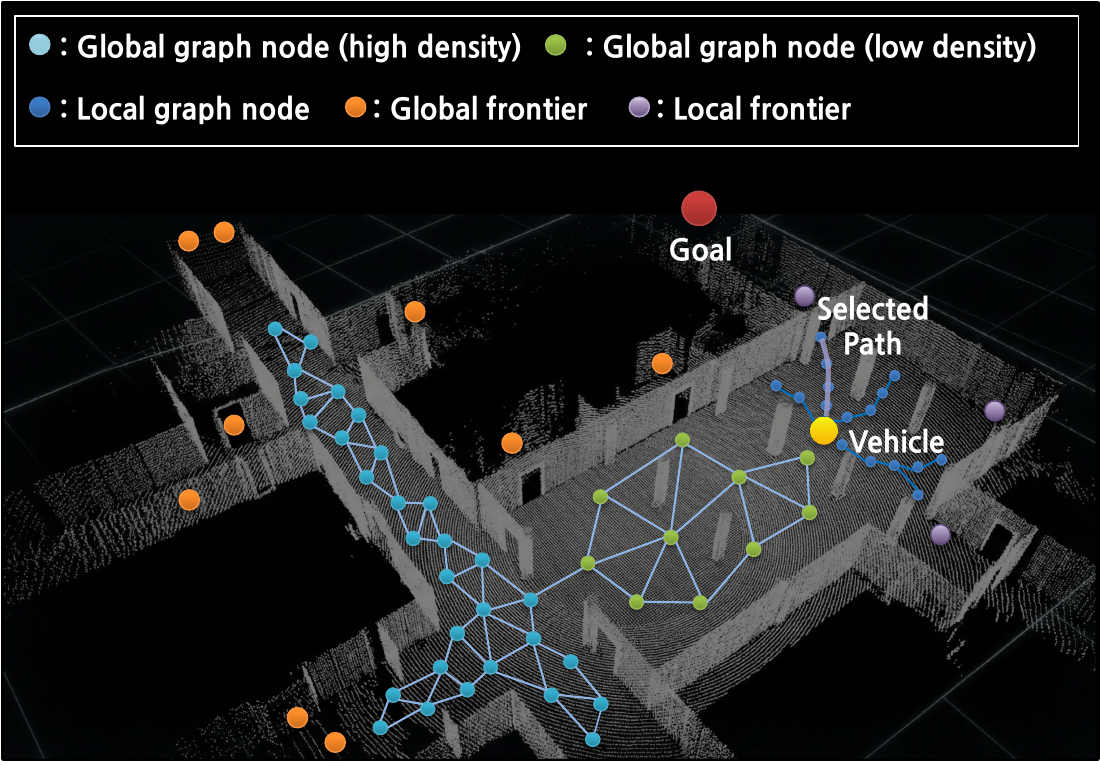}
    \caption{\textbf{Overview of the approach.} Our method navigates the vehicle toward the goal without prior knowledge of the unexplored area. The reactive planning stage in our method generates a path guided by the local frontier within the sensor range. The replanning stage generates a path to the subgoal through the global graph and global frontier. The global graph is generated with low node density in open areas and high node density in narrow areas through adaptive sampling.} 
    \label{fig1}
\end{figure}

In this paper, we propose Frontier-based Path Planning with Adaptive Sampling (FPAS), a novel path generation framework designed for efficient goal-reaching in large-scale, unknown environments (Figure \ref{fig1}). The core insight of FPAS is the reinterpretation of the \textit{frontier} concept, which is traditionally utilized in exploration tasks\cite{zhu2021dsvp} for the goal-reaching problem. Specifically, FPAS leverages frontiers for two primary purposes: guiding navigation and adaptively managing the global graph. The proposed framework operates through two distinct stages. First, the reactive planning stage advances into unobserved regions within the sensor range, where frontiers dictate a forward direction highly advantageous for both goal-reaching and exploration. Second, the replanning stage backtracks from dead-ends or inefficient paths, where frontiers serve as promising subgoals to determine exactly where the robot should retreat for a new search. Furthermore, driven by our key intuition that the required graph density varies with local spatial structure, we introduce an adaptive sampling mechanism based on a local \textit{``openness"} metric, which is directly derived from the spatial characteristics of the frontiers. This mechanism dynamically adjusts node density by employing sparse nodes in open areas to reduce computational burdens, while preserving denser sampling in narrow passages to ensure collision avoidance and connectivity. Ultimately, FPAS sustains high goal-reaching performance while drastically decreasing computational demands.

The main contributions of this paper are threefold. First, we reinterpret the frontier concept to propose a path generation strategy that effectively guides forward progression and selects promising subgoals for backtracking. Second, we introduce an adaptive sampling mechanism based on a frontier-derived openness metric, which dynamically adjusts the global graph's density to significantly alleviate storage and computational burdens. Finally, extensive evaluations demonstrate that FPAS maintains competitive goal-reaching performance while substantially improving computational efficiency over baseline methods.
\section{Related Work}

\subsection{Goal-reaching path planners}

Existing path planning algorithms for goal-reaching can be broadly classified into three main categories: search-based, sampling-based, and visibility-graph-based approaches.

Search-based planners, such as A*\cite{hart1968formal} and D*\cite{stentz1994optimal}, are widely used in path planning. However, standard A* is inefficient in unknown environments because it requires prior knowledge and replans from scratch. To address this, incremental search algorithms like Lifelong Planning A* (LPA*) \cite{koenig2004lifelong}, D*, and its streamlined successor D* Lite \cite{koenig2002d} enable efficient replanning by reusing previous search results. Furthermore, hierarchical approaches \cite{cao2024mpp} enhance efficiency by organizing planning across multiple abstraction levels. Despite these refinements, their fundamental reliance on dense grids or graphs causes search costs to scale poorly in large-scale, unknown environments requiring frequent map updates.

Random sampling-based planners have gained considerable attention in motion planning, particularly for handling high-dimensional and complex environments. Among these, Rapidly-exploring Random Trees (RRTs) are the most prominent, leading to numerous extensions such as bidirectional RRT\cite{kuffner2000rrt}, informed RRT*\cite{gammell2014informed}, and kinodynamic RRT*\cite{webb2012kinodynamic}. To address dynamic or partially known settings, RRT$^X$\cite{otte2016rrtx} introduced a fast replanning approach that adapts to newly recognized information, while recent methods like FHQ-RRT*\cite{dong2025fhqrrt} employ KeyPoints node creation to accelerate high-quality path generation. Despite these advancements, their inherent randomness and goal-directed heuristics often cause unnecessary detours and susceptibility to local minima in complex terrains. Furthermore, the high computational cost of continuously updating the tree restricts their scalability in large-scale unknown environments \cite{devaurs2013parallelizing, jaillet2005adaptive}.

Visibility-graph-based approaches generate paths by connecting mutually visible vertices, a foundational concept rooted in the Probabilistic Roadmap Method (PRM) \cite{kavraki2002probabilistic}. Building upon this, the FAR Planner \cite{yang2022far} has recently emerged as a powerful alternative for goal-reaching in unknown environments, utilizing a polygon-based environment model and dynamic visibility updates. While highly effective, these approaches face two major limitations. First, formalizing irregular obstacles into polygons is often difficult, which decreases the planner's generalizability in unstructured environments. Second, as the graph expands due to complex obstacle distributions or increased exploration time, the cost of checking visibility connections escalates quickly. Consequently, this computational overhead remains a significant bottleneck in large-scale applications.

To overcome the limitations of these prior studies, our proposed FPAS framework aims to appropriately balance goal-reaching with the exploration of unobserved regions, while efficiently managing the memory of environmental information.

\subsection{Next-best-view approaches}

Next-best-view (NBV) approaches are designed to optimize the coverage of an entire environment by quantifying the information that can be obtained from each viewpoint and generating paths that maximize it. Previous research, such as NBVP\cite{bircher2016receding}, employed the RRT algorithm to generate viewpoints and paths. However, this approach focused only on locally expanding the boundaries of the exploration area without considering the overall environment, leading to difficulties in long-term exploration. To overcome this limitation, subsequent studies such as GBP\cite{dang2019graph} and DSVP\cite{zhu2021dsvp} have proposed a dual-stage exploration approach. This approach comprises two stages: one stage focuses on exploring areas within the sensor range, while the other stage involves moving to the explored areas. DSVP, in particular, utilized the boundary between known and unknown areas, known as the frontier, to guide the expansion of RRT. They also proposed a dynamically-expanded RRT that only maintained nodes in free space within the recognition range, thereby reducing computation. Building upon DSVP, we extend its frontier-based strategy to goal-reaching tasks by utilizing frontiers as subgoal candidates. Furthermore, we propose a novel objective function that balances exploration and exploitation, alongside an adaptive sampling method for efficient graph management. Ultimately, these advancements enable highly efficient path generation for goal-reaching in large-scale, unknown environments.
\section{Methodology}

\begin{figure}
    \centering
    \includegraphics[width=0.45\textwidth]{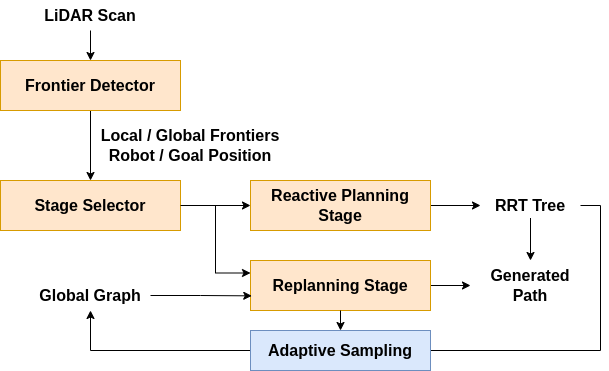}
    \caption{\textbf{The system architecture of FPAS.} The framework consists of four core modules. The frontier detector extracts spatial frontiers to guide navigation. The stage selector dynamically determines when to transition between planning stages. The reactive planning stage generates local paths into unobserved regions within the sensor range. Finally, the replanning stage backtracks from dead-ends and utilizes adaptive sampling to efficiently generate the global graph.}
    \label{fig2}
\end{figure}

In this section, we present the Frontier-based Path Planning with Adaptive Sampling (FPAS) framework. A key idea of FPAS is to repurpose exploration frontiers as goal-reaching guiding signals, enabling both navigation guidance and adaptive management of the global graph. The framework consists of two stages: a reactive planning stage that advances into unobserved regions, and a replanning stage that backtracks to a promising subgoal when encountering dead-ends or inefficient paths. The transition between these two stages is determined by a stage selector. The remainder of this section is organized as follows. Subsection \ref{subsection:stage_selector} describes the stage selector, while Subsections \ref{subsection:reactive_stage} and \ref{subsection:replanning_stage} detail the reactive planning stage and the replanning stage, respectively. Figure \ref{fig2} provides an illustration of the overall system.

\subsection{Stage Selector}
\label{subsection:stage_selector}

The role of the stage selector is to choose whether to use reactive planning or replanning. To make this decision, the stage selector leverages frontier information updated at each time step. First, the frontier detector detects and updates the frontier that borders the free space and unknown areas within the sensor range at each time step. The detected frontier is first classified as a local frontier and then reclassified as a global frontier once it moves out of the sensor range. We denote these as $F^{local}$ and $F^{global}$, respectively. The stage selector determines which stage to enter by comparing $F^{local}$ and $F^{global}$. If there is no $F^{local}$ available, which can occur when there is no further reachable area left to explore (such as in a dead-end situation), the stage selector immediately selects the replanning stage. Otherwise, it selects the local frontier $\hat{F}^{local}$ closest to the goal $g$, and checks whether any global frontier $F^{global}_i$ is closer to $g$ than $\hat{F}^{local}$ . To prioritize the reactive planning stage and prevent frequent switching between stages, we compare $\hat{F}^{local}$ and $F^{global}$ with a margin ($\beta_1, \beta_2 > 0$) for the distance from the vehicle and the distance to the goal, respectively. If any of the $F^{global}_i$ satisfies these conditions, the stage selector chooses the replanning stage. The overall process is described in Algorithm \ref{alg:stage}. 

\begin{algorithm}
    \caption{Stage Selector}
    \label{alg:stage}
    \begin{algorithmic}[1]
        \STATE {{\bfseries Input:} robot position, $g$, $F^{local}$, $F^{global}$}

        \IF{$F^{local}=\emptyset$}
            \STATE \textbf{return} ReplanningStage()
        \ENDIF        
            \STATE $\hat{F}^{local} \gets \argmin_{F^{local}_i\in F^{local}} dist(F^{local}_i, g)$            
        \STATE $SetReplan \gets \FALSE$

        \FOR{$ F^{global}_i \in F^{global}$}
            \IF{$dist(\hat{F}^{local}, robot) + \beta_1 < dist(F^{global}_i , robot) $}
                \IF{$dist(\hat{F}^{local}, g) + \beta_2 > dist(F^{global}_i , g)$}
                    \STATE $SetReplan \gets$ \TRUE
                    \STATE break;
                \ENDIF
            \ENDIF
        \ENDFOR
        \IF{$SetReplan$}
            \STATE \textbf{return} ReplanningStage()
        \ELSE
            \STATE \textbf{return} ReactivePlanningStage()
        \ENDIF
    \end{algorithmic}
\end{algorithm}

\begin{algorithm}
    \caption{ReactivePlanningStage()}
    \label{alg:reactive}
    \begin{algorithmic}[1]
        \STATE {{\bfseries Input:} robot position, $g$, $F^{local}$}
        \STATE $\mathcal{N}\gets\text{DynamicRRT()}$
        \STATE $\mathcal{P} \gets\text{ExtractPaths}(\mathcal{N})$
        \FOR{all $P_i \in \mathcal{P}$}
            \STATE compute $I(P_i)$ using Eq. \ref{eq:Info_path}
            \STATE Compute $GoalDist(P_i)$ using Eq. \ref{eq:goal_dist}
            \STATE Compute $Obj(P_i)$ using Eq. \ref{eq:overall_cost}
        \ENDFOR
    % \STATE Find best path $\hat{P}_i$ using Eq. \ref{eq:bestpath}
    \STATE $\hat{P}_i \gets$ best path using Eq. \ref{eq:bestpath}
    \STATE \textbf{return} $\hat{P}_i$

    \end{algorithmic}
\end{algorithm}

\subsection{Reactive Planning Stage}
\label{subsection:reactive_stage}

In the reactive planning stage, local paths are determined within the sensor range using a dynamically-expanded RRT \cite{zhu2021dsvp}. A tree is iteratively generated and expanded within the local perception field. Initially, the RRT is rooted at the vehicle's current position. From the second iteration onward, the existing tree is preserved while new nodes are added via random sampling within the sensor range, facilitating efficient local navigation. We prioritize sampling near $\hat{F}^{local}$ obtained from the stage selector to increase the efficiency of sampling. During this process, nodes that are too close to the existing tree are excluded, and we use OctoMap\cite{hornung2013octomap} to determine the traversability between nodes. Here, the OctoMap is updated as the environment is explored. The nodes comprising the tree are denoted as $n_i\in \mathcal{N}$. The position of the vehicle is denoted as $n_0$, and we define the path from $n_0$ to $n_i$ within the tree as $P_i\in\mathcal{P}$. Additionally, $n_i^j$ represents the $j$-th node that composes $P_i$.

The path selection balances (i) the amount of new environmental information that can be acquired and (ii) progress toward the final goal $g$. The amount of new environmental information is quantified as an information gain\cite{dang2019graph}, which we denote as $Gain(\cdot)$. This information gain is calculated by determining the number of newly perceived voxels from the node $n_i^j$. The total information gain of a candidate path $P_i$ is then defined as follows:

\begin{equation} \label{eq:Info_path}
    I(P_i)=\sum_{n_i^j\in P_i}Gain(n_i^j).
\end{equation}

The cost of the goal distance $GoalDist(P_i)$ is calculated as the sum of the length of the path and the distance from the last node of the path to the goal.

\begin{equation} \label{eq:goal_dist}
    GoalDist(P_i)=PathLength(P_i)+dist(n_i, g).
\end{equation}

As the vehicle approaches the goal, $GoalDist(P_i)$ decreases, resulting in $I(P_i)$ becoming the dominant factor in the objective function. This may lead to $GoalDist(P_i)$ being disregarded. To balance $GoalDist(P_i)$ and $I(P_i)$, we introduce a sigmoid-based scaling function to dynamically reduce the impact of information gain as the vehicle gets closer to the goal. This function, denoted as $S(x)$, is defined as follows:

\begin{equation} \label{eq:sigmoid}
    S(x)=k_1\cdot \left(\frac{1}{1+e^{-x/k_2}}-\frac{1}{2}\right),
\end{equation}

where $k_1$ and $k_2$ are scaling constants. Combining these terms, we define the final objective as:

\begin{equation} \label{eq:overall_cost}
    Obj(P_i)=I(P_i)\cdot S(dist(n_0,g))-GoalDist(P_i),
\end{equation}

In the reactive planning stage, we consider only paths that provide information gain to prevent the vehicle from moving towards already explored areas. To achieve this, we exclude those with zero information gain from consideration. We select the path with the highest objective value among the paths:

\begin{equation} \label{eq:bestpath}
    \hat{P}_i=\argmax_{P_i\in\mathcal{P}}Obj(P_i) \ \ \text{s.t.}\ I(P_i)\ne0.
\end{equation}

Overall, the reactive planning stage drives the vehicle toward a promising local frontier by selecting information-rich paths while maintaining goal-directed progress. This reactive planning procedure is summarized in Algorithm \ref{alg:reactive}.

\begin{algorithm}
    \caption{ReplanningStage()}
    \label{alg:replan}
    \begin{algorithmic}[1]
        \STATE {{\bfseries Input:} $\mathcal{G}$, $\mathcal{P}$, $robot$, $g$, $F^{local}$, $F^{global}$}
        \STATE $N_{local} \gets |F^{local}|$
        \STATE Compute $Open(N_{local})$ using Eq. \ref{eq:openness}
        \FOR{all $P_i\in\mathcal{P}$}
            \IF{$I(P_i)>0$}
                \FOR{all $n_i^j \in P_i$}
                    \STATE Update $\mathcal{G}$ using adaptive sampling s.t. Eq.  \ref{eq:gg_condition}
                \ENDFOR
            \ENDIF
        \ENDFOR
        \STATE Find best global frontier $\hat{F}^{global}$ using Eq. \ref{eq:bestsubgoal}
        \STATE \textbf{return} $\text{DijkstraPath}(\mathcal{G}, robot, \hat{F}^{global})$
    \end{algorithmic}
\end{algorithm}

\subsection{Replanning Stage with Adaptive Sampling}
\label{subsection:replanning_stage}

\begin{figure}
    \centering
    \begin{subfigure}{0.4\textwidth}
        \centering
        \includegraphics[width=\textwidth]{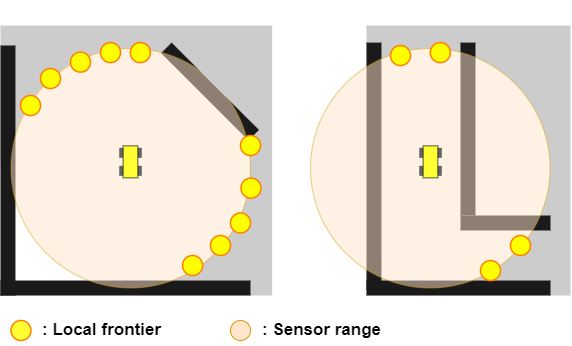}
        \caption{Detected local frontiers}
        \label{fig3a}
    \end{subfigure}
    \vskip 0.5cm
    \begin{subfigure}{0.4\textwidth}
        \centering
        \includegraphics[width=\textwidth]{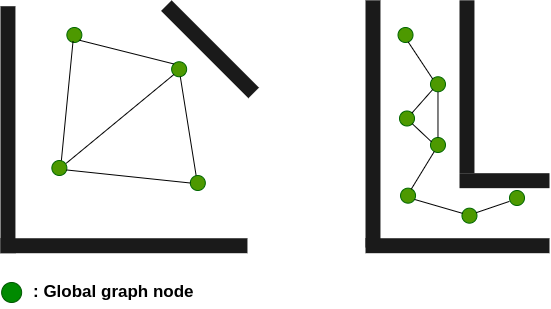}
        \caption{Adaptively sampled global graph}
        \label{fig3b}
    \end{subfigure}
    \caption{(a) Local frontiers (yellow dots) are detected at the boundary between the sensor range (orange circle) and the unknown area. The number of local frontiers increases in open areas (left), while fewer local frontiers are found in narrow passages (right). (b) Openness is measured by the number of local frontiers detected. Nodes from the reactive planning stage are adaptively sampled based on the degree of openness of the area and added to the global graph (green dots).}
    \label{fig3}
\end{figure}

Upon entering the replanning stage, the primary objective is to evaluate and select an appropriate subgoal for backtracking. Specifically, the previously accumulated global frontiers $F^{global}$ are considered as subgoal candidates for a new search. We utilize the following formula to obtain the target subgoal $\hat{F}^{global}$:

\begin{equation} \label{eq:bestsubgoal}
    \begin{aligned}
        \hat{F}^{global}=\argmin_{F^{global}_i \in F^{global}} dist(F^{global}_i,g).
    \end{aligned}
\end{equation}

The environment information acquired during the reactive planning stage is stored as a topological map, denoted as the global graph $\mathcal{G}$. While a straightforward approach would be to retain all nodes generated by the local trees, the continuous accumulation of nodes during exploration inevitably leads to an unbounded increase in graph size. This unbounded growth imposes significant computational burdens on path searching and graph maintenance. To mitigate this bottleneck, we propose an adaptive sampling method that dynamically adjusts the graph's node density, ensuring the efficient construction and long-term management of the global graph. Candidate nodes for updating the global graph are obtained from the reactive planning paths $P_i$ with nonzero information gain ($I(P_i)>0$).  Excluding paths with no information gain prevents oversampling of already explored areas.

We determine the node density based on whether a given region is an open space or a narrow passage. We term this spatial characteristic ``openness" and evaluate it using local frontiers, $F^{local}$. As illustrated in Figure \ref{fig3}, a greater number of local frontiers tends to be detected in expansive open areas compared to confined spaces. Therefore, letting $N_{local}$ denote the number of detected $F^{local}$ and $N_{max}$ denote the maximum possible number of detectable local frontiers, the openness metric can be formulated as follows:

\begin{equation} \label{eq:openness}
    Open(N_{local})=\frac{N_{local}}{N_{max}}.
\end{equation}

$Open(\cdot)$ has a range of 0 to 1, and the closer it is to 1, the more likely the region is to be an open area. We utilize this openness to determine the distance between nodes during the process of updating the global graph and adjust the density of the global graph accordingly. 

Finally, the decision of whether to add candidate nodes to the global graph is made based on their distance from existing nodes in the graph and their traversability. To do this, we first find the node $n_g$ in the global graph that is closest to the node we want to add. Then, we check if the distance between the two nodes satisfies the following condition:

\begin{equation}
    \begin{aligned}\label{eq:gg_condition}
    dist(n^j_i, n_g)>\delta_{min}\cdot (1+Open(N_{local})), \\
dist(n^j_i, n_g)<\delta_{max}\cdot (1+Open(N_{local})),
    \end{aligned}
\end{equation}

where $\delta_{min}$ and $\delta_{max}$ represent the minimum and maximum distance thresholds, respectively. This condition ensures that higher openness can lead to lower density by increasing the distance between nodes in global graph. If these conditions are satisfied, edges are generated between nearby nodes in the global graph through traversability checks. To find the path to reach the subgoal, we first identify the node closest to the subgoal and the node closest to the vehicle within the global graph. Then, we use Dijkstra's algorithm to find the path between these two nodes. The replanning procedure is detailed in Algorithm \ref{alg:replan}. 
\section{Experiments}

Our experiments aim to verify whether the FPAS framework can generate near-optimal paths while minimizing computational costs in large-scale, unknown environments. We designed a sequential multi-goal reaching task to evaluate the balance between exploration and exploitation. This approach explicitly demonstrates the planner's efficiency in maintaining and reusing the global graph over extended periods, which a single-goal scenario cannot adequately capture. To validate FPAS across diverse terrains, we conducted simulations in three environments from the Autonomous Exploration Development Environment \cite{cao2022autonomous}: Forest, Tunnel, and Garage. The Forest environment predominantly consists of open spaces populated with numerous trees and a few small buildings. The Tunnel environment features a maze-like structure composed of narrow passages, which frequently induces dead-ends and heavily tests the planner's replanning capabilities. The Garage environment presents a complex mixture of both open spaces and narrow passages, making it the most suitable setting to evaluate the adaptive density control mechanism, a core feature of FPAS. In all simulations, the vehicle was equipped with a LiDAR sensor configured with a maximum detection range of 30 meters, and the vehicle's speed was set to 2m/s.

\begin{figure}[t]
    \centering
    \includegraphics[width=0.38\textwidth]{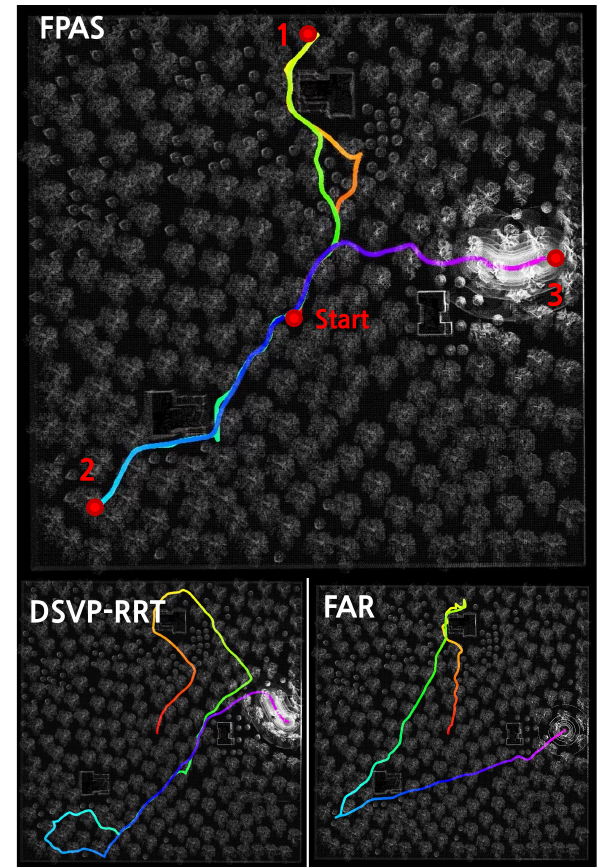}
    \caption{Vehicle trajectory in a forest environment. Color of the trajectory changes according to the sequence of time in which the path was traversed. Three goals are given sequentially.} 
    \label{fig:forest}
\end{figure}
\begin{table}[t]
    \caption{Average processing time and travel distance in forest environment}
    \begin{subtable}[h]{0.47\textwidth}
        \centering
        \begin{tabularx}{\columnwidth}{|c||Y|Y|Y|}
        \hline
        & DSVP-RRT & FAR & FPAS (ours) \\
        \hline \hline
        Goal 1 & 4.61 & 48.14 & \textbf{4.42}\\
        \hline
        Goal 2& 5.73 & 83.75 & \textbf{2.76}\\
        \hline
        Goal 3 & 6.25 & 87.81 & \textbf{2.23}\\
        \hline
        \hline
        Average & 5.53 & 73.23 & \textbf{3.14} \\
        \hline
       \end{tabularx}
       \caption{Average processing time (ms)}
       \label{tab:forest_a}
    \end{subtable}
    \hfill
    \vskip 0.3cm
    \begin{subtable}[h]{0.47\textwidth}
        \centering
        \begin{tabularx}{\columnwidth}{|c||Y|Y|Y|}
        \hline
        & DSVP-RRT & FAR & FPAS (ours)\\
        \hline \hline
        Goal 1 & \textbf{90.32} & 94.52 & 92.17\\
        \hline
        Goal 2 & 229.69 & \textbf{149.07} & 170.77\\
        \hline
        Goal 3 & 197.09 & \textbf{138.53} & 173.75\\
        \hline
        \hline
        Total & 517.1 & \textbf{382.12} & 436.69 \\
        \hline
       \end{tabularx}
        \caption{Travel distance (m)}
        \label{tab:forest_b}
     \end{subtable}
     \label{tab:forest}
     \vspace{-5pt}
\end{table}

To objectively benchmark the performance of FPAS, we compared it against two competitive baselines. The first baseline is a modified Dynamically-expanded RRT. Because the standard RRT is unsuitable for large-scale unknown environments, we adopted the dynamically-expanded RRT structure used in the DSVP algorithm\cite{zhu2021dsvp}. To ensure a controlled comparison, its path selection objective was modified to be identical to that of FPAS. Hereafter, we refer to this baseline as DSVP-RRT. The second baseline is the FAR Planner, a state-of-the-art method for goal-reaching in unknown environments utilizing a visibility graph. We evaluated all methods using travel distance and average processing time (computation time per simulation step). Results are averaged over 10 independent runs per environment. All evaluations were performed on a system equipped with a 5.2GHz Intel Core i9 CPU.

\subsection{Results}

\begin{figure}[t]
    \centering
    \includegraphics[width=0.4\textwidth]{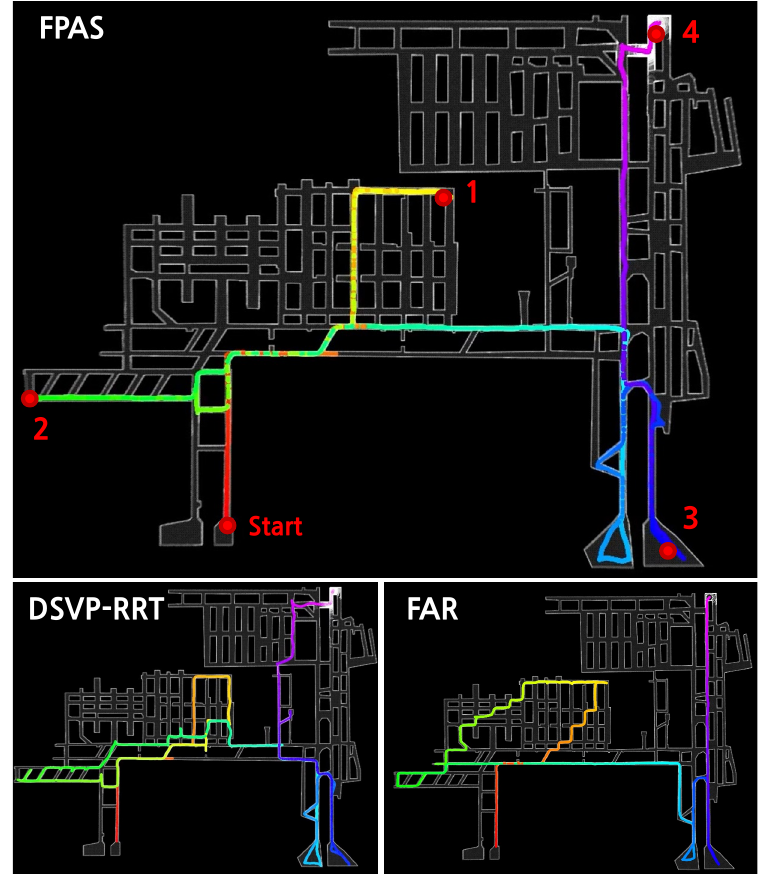}
    \caption{Vehicle trajectory in a tunnel environment. Four goals are given sequentially.} 
    \label{fig:tunnel}
\end{figure}
\begin{table}[t]
    \caption{Average processing time and travel distance in tunnel environment}
    \begin{subtable}[h]{0.47\textwidth}
        \centering
        \begin{tabularx}{\columnwidth}{|c||Y|Y|Y|}
        \hline
        & DSVP-RRT & FAR & FPAS (ours)\\
        \hline \hline
        Goal 1 & 12.31 & \textbf{7.22} & 14.52\\
        \hline
        Goal 2 & 11.42 & \textbf{10.56} & 11.25\\
        \hline
        Goal 3 & 16.05 & \textbf{12.67} & 11.23\\
        \hline
        Goal 4 & 14.50 & \textbf{10.58} & 11.86\\
        \hline
        \hline
        Average & 13.57 & \textbf{10.26} & 12.22 \\
        \hline
       \end{tabularx}
       \caption{Average processing time (ms)}
       \label{tab:tunnel_a}
    \end{subtable}
    \hfill
    \vskip 0.3cm
    \begin{subtable}[h]{0.47\textwidth}
        \centering
        \begin{tabularx}{\columnwidth}{|c||Y|Y|Y|}
        \hline
        & DSVP-RRT & FAR & FPAS (ours)\\
        \hline \hline
        Goal 1 & 259.48 & \textbf{234.89} & 259.65\\
        \hline
        Goal 2& 310.79 & 309.26 & \textbf{306.07}\\
        \hline
        Goal 3 & 686.41 & 644.49 & \textbf{619.38}\\
        \hline
        Goal 4 & 385.87 & \textbf{247.59} & 310.31\\
        \hline
        \hline
        Total & 1642.55 & \textbf{1436.23} & 1495.41 \\
        \hline
       \end{tabularx}
        \caption{Travel distance (m)}
        \label{tab:tunnel_b}
     \end{subtable}
     \label{tab:tunnel}
\end{table}

The results in the forest environment are presented in Figure \ref{fig:forest} and Table \ref{tab:forest}. While FAR achieved the shortest travel distance due to the environment's straight-line feasibility, our method prioritizes acquiring new environmental information during reactive planning, making a slight trade-off in distance. However, FPAS drastically outperformed FAR in computational efficiency. FAR suffered a heavy burden from polygonizing non-standardized obstacles (e.g., trees) and managing an increased number of edges in wide areas. Consequently, FPAS achieved an average processing time of 3.14ms, representing a significant reduction compared to FAR (73.23ms) and DSVP-RRT (5.53ms), effectively validating the efficiency of our adaptive sampling approach.

\begin{figure}[t]
    \centering
    \includegraphics[width=0.4\textwidth]{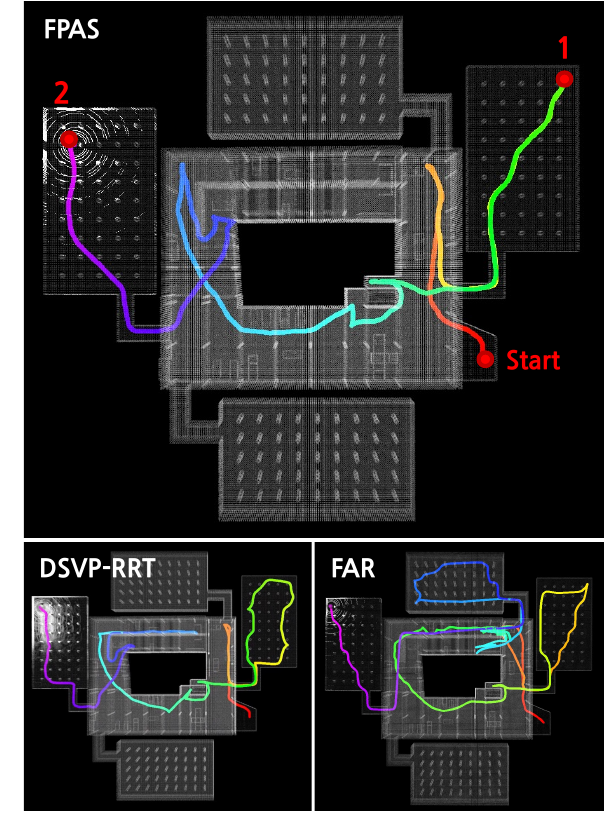}
    \caption{Vehicle trajectory in a garage environment. Two goals are given sequentially.} 
    \label{fig:garage}
\end{figure}
\begin{table}[t]
    \caption{Average processing time and travel distance in garage environment}
    \begin{subtable}[h]{0.47\textwidth}
        \centering
        \begin{tabularx}{\columnwidth}{|c||Y|Y|Y|}
        \hline
        & DSVP-RRT & FAR & FPAS (ours)\\
        \hline \hline
        Goal 1 & 7.57 & 18.49 & \textbf{5.35}\\
        \hline
        Goal 2 & 8.94 & 28.42 & \textbf{4.96}\\
        \hline
        \hline
        Average & 8.26 & 23.46 & \textbf{5.16} \\
        \hline
       \end{tabularx}
       \caption{Average processing time (ms)}
       \label{tab:garage_a}
    \end{subtable}
    \hfill
    \vskip 0.3cm
    \begin{subtable}[h]{0.47\textwidth}
        \centering
        \begin{tabularx}{\columnwidth}{|c||Y|Y|Y|}
        \hline
        & DSVP-RRT & FAR & FPAS (ours)\\
        \hline \hline
        Goal 1 & 175.13 & 177.32 & \textbf{165.52}\\
        \hline
        Goal 2& 470.52 & 811.63 & \textbf{353.43}\\
        \hline
        \hline
        Total & 645.65 & 988.95 & \textbf{518.95} \\
        \hline
       \end{tabularx}
        \caption{Travel distance (m)}
        \label{tab:garage_b}
     \end{subtable}
     \label{tab:garage}
\end{table}

In the tunnel environment (Figure \ref{fig:tunnel}, Table \ref{tab:tunnel}), the maze-like structure necessitates robust replanning and effective use of prior information. Here, FPAS demonstrated superior performance over DSVP-RRT in both processing time (12.22ms vs. 13.57ms) and total travel distance (1495.41m vs. 1642.55m). Unlike the forest, the tunnel's narrow, elongated passages facilitate straightforward visibility graph updates for FAR, yielding a slightly lower processing time (10.26ms) and distance (1436.23m). Nevertheless, FPAS remained highly competitive, proving its robust adaptability and efficiency even in heavily constrained spaces.

Finally, the garage environment (Figure \ref{fig:garage}, Table \ref{tab:garage}) highlights the planner's ability to handle complex, mixed topographies. Similar to the forest, FPAS achieved the lowest average processing time (5.16ms) compared to FAR (23.46ms) and DSVP-RRT (8.26ms). Furthermore, to reach the second goal located inside a room, the baselines failed to utilize a critical narrow passage due to their greedy heuristics, resulting in severe detours. In contrast, FPAS successfully navigated back to the passage via efficient replanning. This led to a substantial reduction in travel distance, with FPAS covering only 518.95m compared to FAR's 988.95m and DSVP-RRT's 645.65m. Ultimately, these evaluations confirm that FPAS maintains consistent, high-speed, and robust performance regardless of varying environmental characteristics.

\subsection{Ablation study}

\begin{figure}[t]
    \centering
    \begin{subfigure}[b]{0.22\textwidth}
        \centering
        \includegraphics[width=\textwidth]{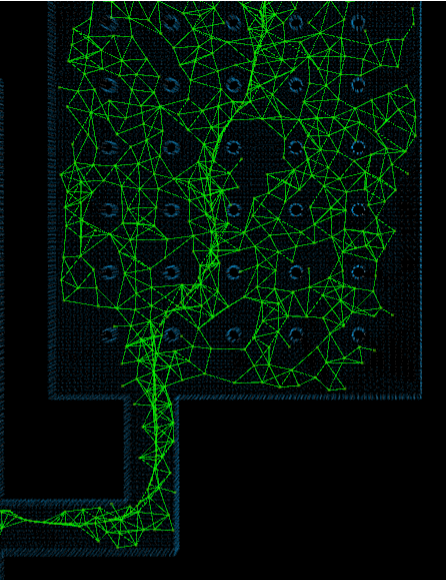}
        \caption{W/o adaptive sampling}
        \label{fig:ablation_a}
    \end{subfigure}
    % \vskip 0.2cm
    \begin{subfigure}[b]{0.22\textwidth}
        \centering
        \includegraphics[width=\textwidth]{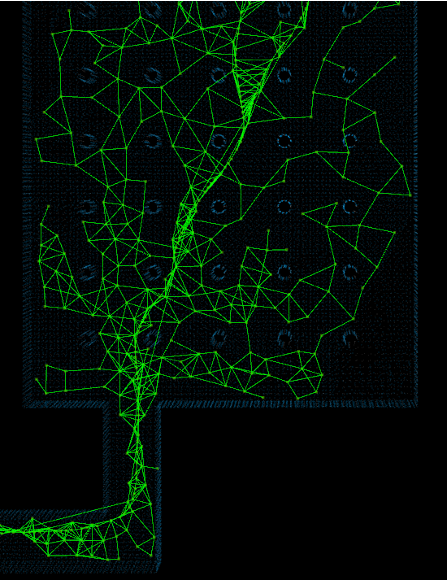}
        \caption{Adaptively sampled}
        \label{fig:ablation_b}
    \end{subfigure}
    \caption{Visualization of the global graph, both without (a) and with (b) applying adaptive sampling, in the garage environment.}
    \label{fig:ablation}
\end{figure}
\begin{table}[t]
    \caption{Ablation study}
        \begin{tabularx}{\columnwidth}{|c||Y|Y|}
        \hline
        & w/o Adaptive Sample & Adaptively Sampled \\
        \hline \hline
        Number of nodes & 2633 & \textbf{1839} \\
        \hline
        Avg. processing time (ms) & 7.01 & \textbf{5.16} \\
        \hline
        Travel dist (m) & \textbf{496.39} & 518.95 \\
        \hline
       \end{tabularx}
     \label{tab:ablation}
\end{table}

An ablation study was conducted in the garage environment to compare the performance improvement of applying adaptive sampling. Figure \ref{fig:ablation} visualizes the global graph according to the application of adaptive sampling. As shown in Figure \ref{fig:ablation_b}, the nodes of the global graph are densely located near the narrow passage located at the bottom, while the top open region has sparsely located nodes. Using adaptive sampling, we were able to reduce the number of nodes required to construct the global graph from 2633 to 1839, leading to a reduction in average processing time from 7.01ms to 5.16ms. However, we observed a slight increase in travel distance from 496.39m to 518.95m when utilizing the adaptively sampled graph. This may be attributed to the fact that the adaptive sampling process used to sparsify the graph can result in the loss of information needed for the shortest path generation. We will leave this issue as future work for our method.
\section{Conclusion}

In this paper, we presented Frontier-based Path Planning with Adaptive Sampling (FPAS), a novel framework designed to achieve efficient goal-reaching in large-scale, unknown environments. By reinterpreting the frontier concept, FPAS effectively balances forward exploration into unobserved regions with optimal backtracking when encountering dead-ends. Furthermore, we introduced an adaptive sampling mechanism based on a frontier-derived openness metric. This approach dynamically adjusts the density of the global graph, significantly alleviating computational and memory burdens by employing sparse nodes in open spaces while maintaining dense sampling in narrow passages.

Despite its advantages, the current framework presents limitations in path optimality and environmental scalability. Deliberate graph sparsification in open areas can cause minor deviations from the absolute shortest paths. Additionally, its reliance on 2D/2.5D representations limits performance in fully 3D environments (e.g., UAVs) or for vehicles with non-holonomic constraints. To address these issues, our future work will refine the adaptive sampling process to enforce minimum path length constraints, ensuring near-optimality alongside computational efficiency. Furthermore, we plan to extend the openness metric to 3D voxel spaces and incorporate kinodynamic constraints into trajectory generation, expanding the framework's applicability to more complex robotic systems.

\addtolength{\textheight}{-1cm}   % This command serves to balance the column lengths
                                  % on the last page of the document manually. It shortens
                                  % the textheight of the last page by a suitable amount.
                                  % This command does not take effect until the next page
                                  % so it should come on the page before the last. Make
                                  % sure that you do not shorten the textheight too much.

%%%%%%%%%%%%%%%%%%%%%%%%%%%%%%%%%%%%%%%%%%%%%%%%%%%%%%%%%%%%%%%%%%%%%%%%%%%%%%%%

%%%%%%%%%%%%%%%%%%%%%%%%%%%%%%%%%%%%%%%%%%%%%%%%%%%%%%%%%%%%%%%%%%%%%%%%%%%%%%%%

%%%%%%%%%%%%%%%%%%%%%%%%%%%%%%%%%%%%%%%%%%%%%%%%%%%%%%%%%%%%%%%%%%%%%%%%%%%%%%%%

%%%%%%%%%%%%%%%%%%%%%%%%%%%%%%%%%%%%%%%%%%%%%%%%%%%%%%%%%%%%%%%%%%%%%%%%%%%%%%%%

% References are important to the reader; therefore, each citation must be complete and correct. If at all possible, references should be commonly available publications.

\bibliographystyle{IEEEtran}
\bibliography{ref}

\end{document}